\documentclass{article}

\usepackage{microtype}
\usepackage{graphicx}
\usepackage{subfigure}
\usepackage{booktabs} % for professional tables

\usepackage{hyperref}

\usepackage[accepted]{icml2025}

\usepackage{amsmath}
\usepackage{amssymb}
\usepackage{mathtools}
\usepackage{amsthm}

\usepackage[capitalize,noabbrev]{cleveref}

\theoremstyle{plain}
\newtheorem{theorem}{Theorem}[section]

\newtheorem{problem}{Problem}[section]
\theoremstyle{definition}
\newtheorem{definition}[theorem]{Definition}

\theoremstyle{remark}

\usepackage{times}
\usepackage{subcaption}

\usepackage[textsize=tiny]{todonotes}

\icmltitlerunning{Unsupervised Structural-Counterfactual Generation under Domain Shift}

\begin{document}

\twocolumn[
\icmltitle{Unsupervised Structural-Counterfactual Generation under Domain Shift}

% It is OKAY to include author information, even for blind
% submissions: the style file will automatically remove it for you
% unless you've provided the [accepted] option to the icml2025
% package.

% List of affiliations: The first argument should be a (short)
% identifier you will use later to specify author affiliations
% Academic affiliations should list Department, University, City, Region, Country
% Industry affiliations should list Company, City, Region, Country

% You can specify symbols, otherwise they are numbered in order.
% Ideally, you should not use this facility. Affiliations will be numbered
% in order of appearance and this is the preferred way.
\icmlsetsymbol{equal}{*}

\begin{icmlauthorlist}
\icmlauthor{Krishn V. Kher}{equal,yyy}
\icmlauthor{Lokesh Badisa}{equal,yyy}
\icmlauthor{Saksham Mittal}{yyy}
\icmlauthor{K.V.D Sri Harsha$^{\dagger}$}{xxx}
\icmlauthor{C.G Sowmya$^{\dagger}$}{xxx}
\icmlauthor{SakethaNath Jagarlapudi}{yyy}
% \icmlauthor{Firstname6 Lastname6}{sch,yyy,comp}
% \icmlauthor{Firstname7 Lastname7}{comp}
% \icmlauthor{}{sch}
% \icmlauthor{Firstname8 Lastname8}{sch}
% \icmlauthor{Firstname8 Lastname8}{yyy,comp}
% \icmlauthor{}{sch}
% \icmlauthor{}{sch}
\end{icmlauthorlist}

% \icmlaffiliation{yyy}{Indian Institute of Technology, Hyderabad}
% \icmlaffiliation{comp}{Company Name, Location, Country}
% \icmlaffiliation{sch}{School of ZZZ, Institute of WWW, Location, Country}

% \icmlcorrespondingauthor{Firstname1 Lastname1}{first1.last1@xxx.edu}
% \icmlcorrespondingauthor{Firstname2 Lastname2}{first2.last2@www.uk}

% You may provide any keywords that you
% find helpful for describing your paper; these are used to populate
% the "keywords" metadata in the PDF but will not be shown in the document
\icmlkeywords{Neural Causal Models, Counterfactual Inference, Domain Adaptation}

\vskip 0.3in
]

% this must go after the closing bracket ] following \twocolumn[ ...

% This command actually creates the footnote in the first column
% listing the affiliations and the copyright notice.
% The command takes one argument, which is text to display at the start of the footnote.
% The \icmlEqualContribution command is standard text for equal contribution.
% Remove it (just {}) if you do not need this facility.

%\printAffiliationsAndNotice{}  % leave blank if no need to mention equal contribution
% \printAffiliationsAndNotice{\icmlEqualContribution} % otherwise use the standard text.

\begin{abstract}
Motivated by the burgeoning interest in cross-domain learning, we present a novel generative modeling challenge: generating counterfactual samples in a target domain based on factual observations from a source domain. Our approach operates within an unsupervised paradigm devoid of parallel or joint datasets, relying exclusively on distinct observational samples and causal graphs for each domain. This setting presents challenges that surpass those of conventional counterfactual generation. Central to our methodology is the disambiguation of exogenous causes into effect-intrinsic and domain-intrinsic categories. This differentiation facilitates the integration of domain-specific causal graphs into a unified joint causal graph via shared effect-intrinsic exogenous variables. We propose leveraging Neural Causal models within this joint framework to enable accurate counterfactual generation under standard identifiability assumptions. Furthermore, we introduce a novel loss function that effectively segregates effect-intrinsic from domain-intrinsic variables during model training. Given a factual observation, our framework combines the posterior distribution of effect-intrinsic variables from the source domain with the prior distribution of domain-intrinsic variables from the target domain to synthesize the desired counterfactuals, adhering to Pearl's causal hierarchy. Intriguingly, when domain shifts are restricted to alterations in causal mechanisms without accompanying covariate shifts, our training regimen parallels the resolution of a conditional optimal transport problem. Empirical evaluations on a synthetic dataset show that our framework generates counterfactuals in the target domain that very closely resemble the ground truth. 
% We also present simulations on benchmark image datasets to visually show that our counterfactuals are meaningful.
\end{abstract}

% Empirical evaluations on synthetic datasets demonstrate that our framework adeptly generates target domain counterfactuals that closely align with the ground truth, underscoring its efficacy and innovation in this emergent generative task.

\section{Introduction}\label{sec:intro}
Counterfactual inference and generation are tasks that have diverse applications in a myriad of fields ranging from biology \cite{10.1007/s10618-021-00818-9, Feuerriegel_2024, QIU2020265} to computer vision \cite{pan2024counterfactual, Yang2020CausalVAEDR, pmlr-v202-de-sousa-ribeiro23a, 9706958} and natural language processing \cite{jin2023cladder, wu2023interpretability, hu2021a, ijcai2024p721}, etc. While some applications employ the term "counterfactual" informally \cite{jeanneret2024text, 52906}, this work adopts the formal definition of structural counterfactuals as per Pearl's hierarchy of causation \cite{10.1145/3501714.3501743, a1ba6a14f6cc486880a9789690c52025}. Current research on counterfactual inference typically confines itself to single domains or fixed causal models, neglecting potential domain shifts or alterations in causal mechanisms. Motivated by the growing interest in cross-domain learning, we address scenarios where the causal model experiences a general domain shift. Specifically, we allow both the causal graph and the underlying causal mechanisms to vary between domains, alongside possible covariate shifts \cite{10.5555/3042817.3043028}. For simplicity, our focus is on the two-domain case, though our framework readily extends to multiple domains. We consider an unsupervised setting where parallel or joint data across domains is unavailable. Instead, observational data and causal graphs are provided separately for each domain. This approach is more practical, as simultaneous observation of cause-effect relationships before and after a domain shift may be infeasible in practice. However, it introduces significant challenges compared to traditional counterfactual generation. To the best of our knowledge, counterfactual generation under such unsupervised domain shifts is an unexplored problem in the literature.

To adequately define structural counterfactuals, a joint causal graph that includes all cause \& effect variables across both domains is required. We construct this joint graph by distinguishing cause variables into two categories: those intrinsic to the effect variable and those intrinsic to the domain. For instance, in image generation, the effect-intrinsic variable could represent the object's identity, while the domain-intrinsic variable might denote the rendering style (e.g., cartoon versus artist-drawn images \cite{Kim2020U-GAT-IT:}). We integrate the two causal graphs by fusing their effect-intrinsic variables (both endogenous and exogenous), as illustrated in Figure~\ref{fig:jointCausalGraph}. Here, subscripts $S$ and $T$ denote the source and target domains, respectively. Exogenous variables are categorized into $C$ (effect-intrinsic) and $N$ (domain-intrinsic), with $X$ representing effect-intrinsic endogenous variables\footnote{For simplicity, endogenous domain-intrinsic variables are assumed absent. If present, they remain separate, similar to $N$.}. This fusion of $C,X$ in the joint causal graph \footnote{Although such fusing of graphs is also done in twin networks \cite{Vlontzos2021EstimatingCC}, both the causal graphs and mechanisms are assumed to be domain-invariant, whereas we allow both to vary.} ensures that the effects in both domains differ solely by their domain-intrinsic variables\footnote{The fused effect-intrinsic variables are chosen to be same as those in the source domain, because the factual is assumed to be on the source side and the counterfactual is expected on the target side. In case of the converse, the fused variables must be chosen as those in the target domain.}. Although we assume identical causal graph structures across domains for clarity of presentation, our methodology can accommodate variations in causal graph structures and interconnects between cause variables.

\begin{figure*}
{%
\subfigure[Separated Causal Models]{%
\label{fig:separate}% label for this sub-figure
\includegraphics[width=0.5\linewidth]{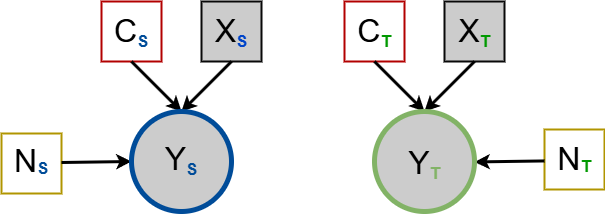}
}\qquad % space out the images a bit
\subfigure[Joint Causal Model] {%
\label{fig:joint}% label for this sub-figure
\includegraphics[width=0.4\linewidth]{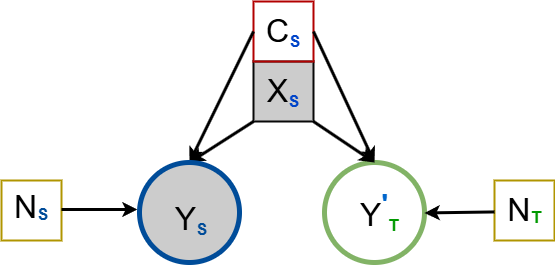}
}
}
{\caption{Illustration explaining the creation of the joint causal graph from individual graphs. Shaded nodes indicate observed variables. $Y^{'}_{\texttt{T}}$ differs from $Y_{\texttt{T}}$ in the prior for $C$: they are abducted from the source, target domains respectively.}}% caption for whole figure
\end{figure*}\label{fig:jointCausalGraph}% label for whole figure
Consequently, we formally define our novel counterfactual generation task as follows: given a factual observation $(x^{\texttt{S}}, y^{\texttt{S}})$ in the source domain, generate counterfactual samples in the target domain, $Y^{\texttt{T}}_{ {x_{\texttt{intv}}}\leftarrow x\ |x^{\texttt{S}},y^{\texttt{S}}}$, where $x_{\texttt{intv}}$ is the intervention value for $x$. Essentially, this involves predicting how the effect $Y$ would appear in the target domain after intervening on the cause $x$ to take the value $x_{\texttt{intv}}$, given the source observation $(x^{\texttt{S}}, y^{\texttt{S}})$. If the causal mechanisms within the joint model are known, counterfactual inference can proceed through Pearl's three-step process \cite{10.5555/1642718}. In practice, the causal mechanisms of the joint graph are unknown and must be learned. We propose using Neural Causal Models (NCMs) \cite{xia2021the}, which are effective for classical counterfactual generation. Training a joint NCM is challenging due to the absence of joint samples. However, since the causal mechanisms remain consistent during sampling, it suffices to learn the causal mechanisms separately for the source and target domains using domain-specific observational data. Accurate identification of exogenous variables is crucial to ensure correct counterfactual inferences. To achieve this, we introduce a novel loss term that facilitates the correct identification of effect-intrinsic variables $C$. Following the learning of a joint NCM, we model the posterior using a neural push-forward generator trained with a Wasserstein-based loss \cite{NIPS2015_a9eb8122, pmlr-v70-arjovsky17a}. The three-step counterfactual generation process is then seamlessly implemented using the learned joint NCM and posterior model.

In scenarios where domain shifts involve only changes in causal mechanisms without covariate shifts, we show that our joint model corresponds with the optimal transport (OT) plan in a conditional OT framework \cite{pmlr-v238-manupriya24a}. This relationship extends existing works that utilize OT for sample generation under domain shifts \cite{korotin2023neural}. To validate our methodology for this novel task, we crafted source and target causal mechanisms adhering to standard identifiability conditions \cite{pmlr-v202-nasr-esfahany23a, nasresfahany2023counterfactualnonidentifiabilitylearnedstructural}. These mechanisms were selected to allow exact computation of the exogenous posterior, thereby precisely determining the counterfactual distribution. Our simulations demonstrate that the proposed methodology correctly estimates the shifted groundtruth counterfactual via a popular two sample test \cite{schrab2021mmd}.

% We then compare the counterfactual, domain-shifted, samples generated from the proposed methodology and those from the ground-truth using a popular two-sample test~\tokrishn{cite}. We empirically observed that most often the two-sample test fails to distinguish the samples, confirming the correctness of our methodology. We also evaluate our methodology in an image generation application, where we found that the samples we produce are indeed very meaningful.

\section{Background}\label{sec:background}
We begin by laying out certain pre-requisites in causality theory (we borrow the notation of \cite{xia2021the} in this regard) and kernels least squares loss that are pertinent to the contributions of this paper, followed by analyzing some prior work that is arguably most closely related to our contributions in this paper. In general we denote random variables by uppercase letters ($W$) and their corresponding values by lowercase letters ($w$). We denote by $\mathcal{D}_{W}$ the domain of $W$ and by $P(W=w)$ the probability of $W$ taking the value $w$ under the probability distribution $P(W)$. We denote by $\Omega(W)$ the domain of values for a random variable $W$. Bold font on a semantic letter indicates a set. 

\subsection{Preliminaries}\label{subsec:prelims}
\textbf{Structural Causal Model} An SCM $\mathcal{M}$ is defined as a tuple $\langle \mathbf{U},\mathbf{V},\mathcal{F},P(\mathbf{U})\rangle$, where $\mathbf{U}$ is a set of exogenous variables with distribution $P(\mathbf{U})$, $\mathbf{V}$ the endogenous variables, and $\mathcal{F}\equiv\{f_1,\ldots,f_n\}$ is a set of functions/mechanisms, where each $f_i:\mathbf{U}_i\times \mathbf{A}_i\rightarrow V_i, \mathbf{U}_i\subset \mathbf{U}, \mathbf{A}_i\subset \mathbf{V}\setminus\{V_i\}$. In simple words, $\mathbf{V}$ are the observed variables, $\mathbf{U}$ are the unobserved variables, and each mechanism ($f_i$) takes as input some subset of exogenous ($\mathbf{U}_i$) and endogenous variables ($\mathbf{A}_i$) and causes/outputs the corresponding observed variable ($V_i$). 
% So $\mathbf{U}_i,\mathbf{A}_i$ are the causes for the effect $V_i$. 
Given a \emph{recursive} SCM, a directed graph is readily induced, in which the nodes correspond to the endogenous variables $\mathbf{V}$, while the directed edges correspond to the causal mechanism from each variable in $\mathbf{A}_i$ to $V_i$. It is assumed that this directed graph is acyclic and as such, $\mathbf{A}_i$ are the parents of $V_i$ in this DAG. Every causal diagram is associated with a set of $\mathcal{L}_1$ constraints, which is a set of conditional independences, $\mathcal{L}_2$ constraints, popularly known as the do-calculus rules for interventions and finally, $\mathcal{L}_3$ constraints, rules that the counterfactual distributions satisfy. Every counterfactual distribution induced by the SCM satisfies all three levels of constraints. Using these 3-layered rules, various statistical, interventional, and counterfactual inferences can be performed in a systematic way. Notably, distributions in the lower levels of the causal hierarchy may not satisfy constraints of higher levels.
Since in most applications the cause-effect relations are known rather than the SCM itself, one interesting modeling question is, given a causal graph, can we come up with a convenient model that is consistent with all the three levels of constraints induced by the graph?
NCMs are one such example of a convenient family of causal models. 
\begin{definition}\label{defn:ncm}
\textit{NCM}, Defn.$2$ in \cite{Xia_Bareinboim_2024} 
Given a causal diagram $\mathcal{G}$, a $\mathcal{G}$-constrained NCM $\widehat{\mathcal{M}_{\theta}}$ over $\mathbf{V}$ with parameters $\mathbf{\theta} = \{\theta_{V_i} : V_i \in \mathbf{V}\}$ is an SCM $\langle \widehat{\mathbf{U}},\widehat{\mathbf{V}},\widehat{\mathcal{F}},P(\mathbf{\widehat{U}})\rangle$ such that (1) $\mathbf{\widehat{U}} =
\{\widehat{U}_\textbf{C} : \textbf{C} \in \mathbb{C}(\mathcal{G})\}$, where
$\mathbb{C}(\mathcal{G})$ is the set of all maximal cliques over bi-directional edges of
$\mathcal{G}$; (2) $\mathcal{\widehat{F}} = \{\widehat{f_{V_i}} : V_i \in \mathbf{V}\}$, where each $\widehat{f_{V_i}}$ is a feedforward neural net
parameterized by $\theta_{V_i} \in \mathbf{\theta}$ mapping $\mathbf{U}_{V_i} \cup \mathbf{A}_{V_i}$ to $V_i$ for $\mathbf{U}_{V_i} = \{
\widehat{U}_{\mathbf{C}} \in \widehat{\mathbf{U}} : V_i \in \mathbf{C}\}$ and $\mathbf{A}_{V_i}
= A_{\mathcal{G}}(V_i)$; (3) $\texttt{Unif}(0,1) \mapsto P(\widehat{U}), \forall~\widehat{U} \in \widehat{\mathbf{U}}$.
\end{definition}
Note that point $3$ in Definition \ref{defn:ncm} above critically lies on the fact that there always exists a neural network that can transform $\texttt{Unif}(0,1)$ to an arbitrary distribution $P$ (cf. Lemma $5$ in \cite{xia2021the}). To facilitate easier learning of NCMs, we assume $\mathcal{N}(0,1) \mapsto P(\widehat{U}), \forall~\widehat{U} \in \widehat{\mathbf{U}}$ in this work. Note that Lemma $5$ can be trivially extended to this case.

\textbf{Kernel Least Squares} It is well known that $\mathbb{E}[Y|X]=\textup{argmin}_{f}\ \mathbb{E}\left[\left\|f(X)-Y\right\|^2\right]$ \cite{brockwell1991time}. When the kernel embeddings of $Y$ are used instead, it is known as kernels least squares, written as: $\mathbb{E}[\Phi(Y)|X]= \textup{argmin}_{f}\ \mathbb{E}\left[\left\|f(X)-\Phi(Y)\right\|^2\right]$, where $\Phi$ is the canonical feature map corresponding to a kernel. In case the kernel is characteristic \cite{JMLR:v12:sriperumbudur11a}, $Y\mapsto\mathbb{E}[\Phi(Y)|X]$ is injective and characterizes the distribution of $Y$. Common examples of characteristic kernels include the radial basis function (\emph{RBF}) kernel and inverse multi-quadric kernel (\emph{IMQ}) kernel among others. Thus, kernel least- squares loss is well-suited for learning conditional distributions (e.g., see \cite{pmlr-v238-manupriya24a} which provides empirical and theoretical results in this regard). We espouse the following derivation from \cite{pmlr-v238-manupriya24a}, which shows how to learn a conditional generator using joint samples, without having to resort to Monte Carlo methods. Let $\mathcal{D}$ be a given dataset of samples drawn from the joint distribution of random variables $\mathrm{P}, \mathrm{Q}$ and let  $\pi_{\mathrm{Q}|\mathrm{P}}^{\gamma}$ be a parametrized (by $\gamma$) conditional generator that we wish to learn. Accordingly, we wish $\pi_{\mathrm{Q}|\mathrm{P}}^{\gamma}(\cdot|p) = \mathrm{s}_{\mathrm{Q}|\mathrm{P}}(\cdot|p), \forall p \in \Omega(\mathrm{P})$. Utilizing the injectivity of a characteristic kernel $\Phi$, we can equivalently rewrite the desired condition as $\int_{\Omega(\mathrm{P})}\Big\|\mathbb{E}_{\pi_{\mathrm{Q}|\mathrm{P}}^{\gamma}(\cdot|p)}[\Phi(Y)] - \mathbb{E}_{\mathrm{s}_{\mathrm{Q}|\mathrm{P}}(\cdot|p)}[\Phi(Y)]\Big\|_2^2\texttt{d}\mathrm{s}_{\mathrm{P}}(p) = 0$. The kernels least squares loss inside the above integral is commonly known as the squared Maximum Mean Discrepancy error, aka $\texttt{MMD}^2$. For the rest of this paper, we take $\Phi$ to be the \emph{IMQ} kernel defined as $k(x, y) = \frac{1}{\sqrt{\varrho + ||x-y||_2^2}} \forall x, y \in \mathbb{R}^d$, where $\varrho$ is a non-negative hyperparameter. This is because we usually observe good results with this kernel. 
Next, they apply a standard result kernel mean embeddings, \cite{Muandet_2017}, which states that $\mathbb{E}[\|G - h(\mathrm{P})\|^2] = \mathbb{E}[\|G - \mathbb{E}[G|\mathrm{P}]\|^2] + \mathbb{E}[\|\mathbb{E}[G|\mathrm{P}]-\mathbb{E}[h(\mathrm{P})]\|^2]$, when $G$ is the kernel mean embedding of $\delta_{\mathrm{Q}}$ and $h(\mathrm{P})$ the kernel mean embedding of $\pi_{\mathrm{Q}|\mathrm{P}}^{\gamma}(\cdot|\mathrm{P})$. This helps us simplify the integral over the marginal of $\mathrm{P}$ in terms of a marginal over the joint distribution of $\mathrm{P}, \mathrm{Q}$, since $\int_{\Omega(\mathrm{P})}\texttt{MMD}^2(\pi_{\mathrm{Q}|\mathrm{P}}^{\gamma}(\cdot|p), \mathrm{s}_{\mathrm{Q}|\mathrm{P}}(\cdot|p))\texttt{d}\mathrm{s}_{\mathrm{P}}(p) + \vartheta(\mathrm{s}) = \int_{\Omega(\mathrm{P})\times\Omega(\mathrm{Q})}\texttt{MMD}^2(\pi_{\mathrm{Q}|\mathrm{P}}^{\gamma}(\cdot|p), \delta_{\mathrm{Q}})\texttt{d}\mathrm{s}_{\mathrm{P},\mathrm{Q}}(p,q)$, where $\vartheta(\mathrm{s}) \geq 0$ is purely a function of the dataset and therefore, does not affect the minima of the right hand side of the equation above. The left hand side integral can be readily estimated empirically, as $\frac{1}{|\mathcal{D}|}\sum_{i=1}^{|\mathcal{D}|}\big\|\frac{1}{\kappa}\sum_{\widetilde{q_i} \sim \pi_{\mathrm{Q}|\mathrm{P}}^{\gamma}(\cdot|p_i)}^{\kappa}\Phi(\widetilde{q_i}) - \Phi(q_i)\big\|^2$. Training the conditional generator over the dataset $\mathcal{D}$ thereby is tantamount to minimizing the previously mentioned empirical estimate with respect to the parameters, $\gamma$. This "trick" is repeatedly used in most of our losses.
\subsection{Prior Work}\label{subsec:priorwork}
\textbf{Disentanglement} Despite the vast amount of literature in disentanglement particularly in the context of causal representation learning \cite{}, there are only a few that are close to the exact setting we assume in this work. \cite{10.5555/3540261.3541519} also studied causally disambiguating domain intrinsic exogenous variables ("style") and an effect intrinsic variables ("context"), but they assume the dataset for either domain to be sampled using the same context distribution, whereas we allow for different observed context distributions in the observational data (cf. Figure \ref{fig:jointCausalGraph}). \cite{pmlr-v206-morioka23a, daunhawer2023identifiability} extended the above work and theoretically analyzed the above setting, known as "multimodal" SCMs. However, they all seem to assume access to paired data across domains that share the same context, which we don't. Additionally, our kernel-based losses critically differ from their contrastive loss based approaches. Lastly, our setting for unpaired observed data is closest to the Out-Of-Variable generalization setting proposed in \cite{guo2024outofvariable}, but we specifically solve (counterfactual) generative tasks in this setting, whereas they restrict themselves to discriminative tasks.

\textbf{Domain Shifts} Informally known by various names in the literature such as Domain Adaptation (DA) (\cite{10.1007/978-3-030-71704-9_65, article}), Translation (\cite{8100115,9010872}), etc. The underlying idea is to train a model using a samples from one distribution, i.e. the source, and be able to predict objects belonging to another distribution, termed as the target distribution. Whilst a plethora of such works exist in this area, to the best of our knowledge, it appears that a causal angle to this problem has largely been overlooked. \cite{NEURIPS2018_39e98420} proposes a causal DAG modeling domain adaptation and incorporates a context variable, as we also do. \cite{10.5555/3524938.3525815} on the other hand delves into this approach deeper and attempts to model the DA problem via structural equation models, but relies on the assumption of having the data generative mechanism to be invariant across domains. Our methodology however, admits different priors as well as mechanisms across domains. 
Lastly, \cite{Zhang_Gong_Schoelkopf_2015} also examine the DA problem in a simple two-variable setting and demonstrate conditions for transfer of the relevant context across domains by considering the appropriate prior and posterior distributions of the target domain. However, they constrain the causal mechanism to linear structural equations. Furthermore, none of the above works involve answering counterfactual queries in the target domain, nor deal with counterfactual generation in its rigorous form. 
\section{Problem Definition and Proposed Methodology}
Consider the joint model presented in Figure \ref{fig:jointCausalGraph}. $Y_{\texttt{S}}$ and $Y_{\texttt{T}}$ represent the effect variables in the source and target domains respectively. Each of these is causally influenced by a common set of effect-intrinsic attributes: $X$ (endogenous), and $C$ (exogenous). The exogenous domain-intrinsic variables $N_{\texttt{S}}$ and $N_{\texttt{T}}$ are assumed to be independent of each other and need not have the same distribution. In other words, the effect variable has the same effect-intrinsic causes both before and after the domain shift. The effect variables may hence only differ in their domain-intrinsic causes, which is meaningful. Overall, the model is assumed to be Markovian to avoid non-identifiability issues (\cite{nasresfahany2023counterfactualnonidentifiabilitylearnedstructural, pmlr-v202-nasr-esfahany23a}). Other such assumptions leading to identifiability are also fine. As noted earlier, the causal mechanism generating the effect variables may change with the domain shift. Thus, we denote $Y_{\texttt{S}}\equiv \mathcal{M}^{Y_\texttt{S}*}(X_{\texttt{S}},C_{\texttt{S}},N_{\texttt{S}})$ and $Y_{\texttt{T}}\equiv \mathcal{M}^{Y_\texttt{T}*}(X_{\texttt{T}},C_{\texttt{T}},N_{\texttt{T}})$, following the given source and target causal graphs. We now formally define our novel generation problem:
\begin{problem}[\textbf{Counterfactual Translation}]
    Let $(x, y)$ be a given sample from the joint distribution $\mathcal{P}(X_{\texttt{S}}, Y_{\texttt{S}})$ induced by the source causal mechanism $\mathcal{M}^{Y_\texttt{S}*}(\cdot, \cdot, \cdot)$. Let $C_{\texttt{S}|(x,y)}$ denote a random variable obeying the posterior distribution induced over $C_{\texttt{S}}$, by conditioning on $(x, y)$ in the source mechanism $\mathcal{M}^{Y_\texttt{S}*}(\cdot, \cdot, \cdot)$. Given a value $x_{\text{intv}}$, the task is to generate samples corresponding to the distribution induced by $\mathcal{M}^{Y_\texttt{T}*}(x_{\text{intv}}, C_{\texttt{S}|(x,y)}, N_{\texttt{T}})$.
\end{problem} 
To motivate this problem, consider the task of person re-identification in surveillance systems, which is crucial for tracking individuals across different camera views in public spaces such as airports or train stations (\cite{9336268}).
For the sake of this illustration, we extend this to a case where a suspect needs to be tracked for migration between tourist locations of two different countries with starkly clothing traditions. 
Imagine that we have two sets of images: one from images of people in region $\texttt{S}$ and the other from those in region $\texttt{D}$, each of them with their native attires respectively. These images may not overlap, and we do not have paired samples showing the same individual captured in both locations. The task is to generate highly probable images of the suspect in the clothing style native to region $\texttt{D}$, given the information that the suspect has blended themselves with the local clothing culture. Furthermore, due to the passage of time and local weather conditions in region $\texttt{D}$, we may assume that certain attributes such as the age, skin color, etc. have changed, that are also provided as additional information.
Incorporating this additional information is however, non-trivial. With no paired training data linking the source and target images, one needs to generate a plausible counterfactual version of the person in the source image, conditioned on images representing the person in region $\texttt{S}$, along with interventions indicating a change in appearance such as age or skin color, and then translate this counterfactual image to the style of clothing adopted in region $\texttt{D}$.

We can now elucidate correspondences of different variables of interest in the formal definition of the problem described above to that setting. $X_{\texttt{S}}$ models intervenable attributes such as 'age', 'skin color', etc. and obeys the distribution of such attributes as observed in region $\texttt{S}$. Analogously, $X_{\texttt{T}}$ models the same in region $\texttt{T}$. While the attributes semantically allude to the same concepts, the distributions of these variables can differ significantly across regions due to varied reasons such as geography, climate, local culture, etc. $C_{\texttt{S}}$ models latent variables that roughly capture the identity of the person, such as their silhouette, posture, etc. These are not explicitly annotated for in the dataset, yet need to be captured to produce an accurate counterfactually translated image of the suspect. Thus conditioning on the known age/skin color of the person and their image in region $\texttt{S}$ would tantamount to inducing a posterior distribution of silhouettes/postures particularly identifying the suspect. Finally, the domain-intrinsic exogenous variables $N_{\texttt{S}}, ~N_{\texttt{T}}$ may capture other unobserved variables such as the clothing culture in the respective regions etc.% Thus, this exogenous noise models what is commonly referred to as 'style' in many image-to-image-based translation works \tokrishn{cite}\cite{}. We emphasize that to the best of our knowledge, a majority of works in this domain do not clearly distinguish between having a common context between two images and having a style that is native to each domain and usually end up mixing the two. Our modelling seeks to clear this ambiguity and provide a rigorous framework to clearly evaluate the effect of conditioning/intervening with each of these causal variables.

\subsection{Modeling and Training}\label{sec:method}
We propose modeling causal mechanisms of interest using NCMs, because of their versatility and competence in executing counterfactually generative tasks \cite{xia2023neural}.
We model the true source and target causal mechanisms $\mathcal{M}^{Y_\texttt{S}*},\mathcal{M}^{Y_\texttt{T}*}$ using neural networks $\widehat{\mathcal{M}^{Y_\texttt{S}}_{\theta_{\texttt{S}}}},\widehat{\mathcal{M}^{Y_\texttt{T}}_{\theta_{\texttt{T}}}}$ with parameters $\theta_{\texttt{S}}, \theta_{\texttt{T}}$ respectively. However, we lack explicit supervision for the exogenous variables, $C_{\texttt{S}/\texttt{T}}, N_{\texttt{S}/\texttt{T}}$.  
Therefore, we adopt a trick promulgated by \cite{xia2021the}, wherein we technically model wherein $\widehat{\mathcal{M}^{Y_\texttt{S}}_{\theta_{\texttt{S}}}},\widehat{\mathcal{M}^{Y_\texttt{T}}_{\theta_{\texttt{T}}}}$ attempt to align with the equivalent (groundtruth) causal models $\widehat{\mathcal{M}^{Y_\texttt{S}*}},\widehat{\mathcal{M}^{Y_\texttt{T}*}}$, where $\widehat{C_{\texttt{S}/\texttt{T}}}, \widehat{N_{\texttt{S}/\texttt{T}}} \sim \mathcal{N}(0,\texttt{I}_{d\times d})$.

Furthermore, we propose modeling the exogenous variables $C_{\texttt{S}/\texttt{T}}, N_{\texttt{S}/\texttt{T}}$, using push-forward neural generators, that conjoined with $\widehat{\mathcal{M}^{Y_\texttt{S}}_{\theta_{\texttt{S}}}},\widehat{\mathcal{M}^{Y_\texttt{T}}_{\theta_{\texttt{T}}}}$, would mimic the structural effect of the exogenous variables on the output causal variables $Y_{\texttt{S}/\texttt{T}}$. Specifically, let $C_{\texttt{S}}\equiv \widehat{m^{C_{\texttt{S}}\#}_{\phi_{C_{\texttt{S}}}}}(\eta^{C_{\texttt{S}}}), C_{\texttt{T}}\equiv \widehat{m^{C_{\texttt{T}}\#}_{\phi_{C_{\texttt{T}}}}}(\eta^{C_{\texttt{T}}})$, where $\eta^{C_{\texttt{S}}},\eta^{C_{\texttt{T}}}$ are samples from a noise distribution that is easy to sample from, e.g. Gaussian. Similarly, let $N_{\texttt{S}}\equiv \widehat{m^{N_{\texttt{S}}\#}_{\zeta_{N_{\texttt{S}}}}}(\eta^{N_{\texttt{S}}}), N_{\texttt{T}}\equiv \widehat{m^{N_{\texttt{T}}\#}_{\zeta_{N_{\texttt{T}}}}}(\eta^{N_{\texttt{T}}})$. Note that such careful modeling of the exogenous variables is needed for two reasons: (i) we wish to disambiguate between the effect-intrinsic and domain-intrinsic variables (ii) the distribution of $C_{\texttt{S}}$ need not be same as that of $C_\texttt{T}$. Similarly, the distribution of $N_{\texttt{S}}$ need not be same as that of $N_{\texttt{T}}$. Classical counterfactual models need not resort to such careful modeling and can safely assume $\widehat{m^{C_{\texttt{S/T}}\#}_{\phi_{C_{\texttt{S/T}}}}}(\eta^{C_{\texttt{S/T}}}),\widehat{m^{N_{\texttt{S/T}}\#}_{\zeta_{N_{\texttt{S/T}}}}}(\eta^{N_{\texttt{S/T}}})$ are some hidden sub-networks in $\widehat{\mathcal{M}^{Y_\texttt{S}}_{\theta_{\texttt{S}}}},\widehat{\mathcal{M}^{Y_\texttt{T}}_{\theta_{\texttt{T}}}}$ itself.

The main reason why training the described joint model is challenging is because it is not practical to assume that the joint samples are available. As mentioned earlier, we make a more practical assumption that cause-effect samples from the source and target models are separately available. Consequently, let $(x^{\texttt{S}}_1,y^{\texttt{S}}_1),\ldots,(x^{\texttt{S}}_{m_1},y^{\texttt{S}}_{m_1})$ denote the observational data in the source domain and let $(x^{\texttt{T}}_1,y^{\texttt{T}}_1),\ldots,(x^{\texttt{T}}_{m_2},y^{\texttt{T}}_{m_2})$ denote the observational data in the target domain. Our losses significantly differ from \cite{xia2021the} in that we mostly train our models via a kernels least squares loss between samples of appropriate distributions whereas they train using negative log-likelihood. Since the causal mechanisms are independent of the sampling procedure, the mechanisms for generating $Y_{\texttt{S}},Y_{\texttt{T}}$ in the source, target-specific NCMs and in the joint NCM are the same. Hence, the source and target observational data can be used to train the networks in the joint model to be $\mathcal{L}_1$ consistent with the true mechanisms. For instance, on the source side the goal is to learn $\widehat{\mathcal{M}^{Y_{\texttt{S}}}_{\theta_{\texttt{S}}}}(X_{\texttt{S}}, \cdot, \cdot)$ such that $P_{\widehat{\mathcal{M}^{Y_{\texttt{S}}}_{\theta_{\texttt{S}}}}}\left(X_{\texttt{S}}, \cdot, \cdot\right) = P_{\widehat{\mathcal{M}^{Y_{\texttt{S}*}}_{\theta_{\texttt{S}}}}}\left(X_{\texttt{S}}, \cdot, \cdot\right)$. To this end, we critically employ the fact that any characteristic kernel $\Phi$ induces an injective map over distributions, stated in Section \ref{subsec:prelims}, as follows:
% \vspace{-0.8cm}
\begin{equation}
\begin{aligned}[t]
\label{eqn:gen_loss_derivation}
P_{\widehat{\mathcal{M}^{Y_{\texttt{S}}}_{\theta_{\texttt{S}}}}}\left(X_{\texttt{S}}, \cdot, \cdot\right) = P_{\widehat{\mathcal{M}^{Y_{\texttt{S}*}}}}\left(X_{\texttt{S}}, \cdot, \cdot\right) \Leftrightarrow P_{\widehat{\mathcal{M}^{Y_{\texttt{S}}}_{\theta_{\texttt{S}}}}}\left(X_{\texttt{S}}=x, \cdot, \cdot\right) \\ = P_{\widehat{\mathcal{M}^{Y_{\texttt{S}*}}}}\left(X_{\texttt{S}}=x, \cdot, \cdot\right),~\forall x \in \Omega(X_{\texttt{S}}) \Leftrightarrow ~\forall x \in \Omega(X_{\texttt{S}}), \\ \mathbb{E}_{Y_{\texttt{S}} \sim P_{\widehat{\mathcal{M}^{Y_{\texttt{S}}}_{\theta_{\texttt{S}}}}}\left(X_{\texttt{S}}=x, \cdot, \cdot\right)}[\Phi(Y_{\texttt{S}})] = \mathbb{E}_{Y_{\texttt{S}} \sim P_{\widehat{\mathcal{M}^{Y_{\texttt{S}*}}}}\left(X_{\texttt{S}}=x, \cdot, \cdot\right)}[\Phi(Y_{\texttt{S}})].
\end{aligned}
\end{equation}
Consequently, we aim to learn a conditional generator $\widehat{\mathcal{M}^{Y_{\texttt{S}}}_{\theta_{\texttt{S}}}}$, using joint samples $\{(x^{\texttt{S}}_i, y^{\texttt{S}}_i)\}_{i=1}^{n}$. This is our preferred loss to train over observational data due to its demonstrated superior performance compared to other traditional losses such as adversarial/KL/Wasserstein losses \cite{pmlr-v238-manupriya24a}. In particular, we use the $\texttt{MMD}^2$ loss between the sample kernel means of the conditional distributions $P_{\widehat{\mathcal{M}^{Y_{\texttt{S}}}_{\theta_{\texttt{S}}}}}(\widehat{C_{\texttt{S}}}, \widehat{N_{\texttt{S}}}|X_{\texttt{S}}=x)$ and $P_{\widehat{\mathcal{M}^{Y_{\texttt{S}}*}}}(\widehat{C_{\texttt{S}}}, \widehat{N_{\texttt{S}}}|X_{\texttt{S}}=x)$ as follows:
% \vspace{-1cm}
\begin{equation}\label{eqn:gen_loss_S}
\begin{aligned}[t]
    \ell^{\texttt{S}}_{\texttt{gen}}(\theta_{\texttt{S}}, \phi_{C_\texttt{S}}, \zeta_{N_{\texttt{S}}}) = \frac{1}{n^{\texttt{S}}_{\texttt{gen}}}\sum_{i=1}^{n^{\texttt{S}}_{\texttt{gen}}}\Bigg\|\Phi(y^{\texttt{S}}_i)-\\\frac{1}{q^{\texttt{S}}_{\texttt{gen}}}\sum_{j=1}^{q^{\texttt{S}}_{\texttt{gen}}}\Phi\left(\widehat{\mathcal{M}^{Y_{\texttt{S}}}_{\theta_{\texttt{S}}}}\left(x^{\texttt{S}}_i, \widehat{m^{C_{\texttt{S}}\#}_{\phi_{C_{\texttt{S}}}}}\left(\eta^{C_{\texttt{S}}}_{ij}\right), \widehat{m^{N_{\texttt{S}}\#}_{\zeta_{N_{\texttt{S}}}}}\left(\eta^{N_{\texttt{S}}}_{ij}\right)\right)\right)\Bigg\|_2^2.
\end{aligned}
\end{equation}
Here, $n^{\texttt{S}}_{\texttt{gen}}$ is the number of samples within the training set (or within a mini-batch, if using an optimization algorithm such as mini-batch SGD) and $q^{\texttt{S}}_{\texttt{gen}}$ the number of noise samples $\eta^{C_{\texttt{S}}/N_{\texttt{S}}}_{ij}$ sampled from $\mathcal{N}(0,I_{d \times d})$ per data point $(x^{\texttt{S}}_i, y^{\texttt{S}}_i)$, used to empirically approximate the kernel mean of $Y_{\texttt{S}}$ when in turn sampled from $P_{\widehat{\mathcal{M}^{Y_{\texttt{S}}}_{\theta_{\texttt{S}}}}}(\widehat{C_{\texttt{S}}}, \widehat{N_{\texttt{S}}}|X_{\texttt{S}}=x)$. For the sake of completeness, we also note that alternatives from conditional generative adversarial networks based literature may also be viable to learn this conditional generator, as we do not claim the above loss as a novel contribution of our work. A similar loss is also introduced to train the target mechanisms as shown below:
\begin{equation}\label{eqn:gen_loss_T}
\begin{aligned}[t]
        \ell^{\texttt{T}}_{\texttt{gen}}(\theta_{\texttt{T}}, \phi_{C_\texttt{T}}, \zeta_{N_{\texttt{T}}}) = \frac{1}{n^{\texttt{T}}_{\texttt{gen}}}\sum_{i=1}^{n^{\texttt{T}}_{\texttt{gen}}}\Bigg\|\Phi(y^{\texttt{T}}_i) - \\ \frac{1}{q^{\texttt{T}}_{\texttt{gen}}}\sum_{j=1}^{q^{\texttt{T}}_{\texttt{gen}}}\Phi\left(\widehat{\mathcal{M}^{Y_{\texttt{T}}}_{\theta_{\texttt{T}}}}\left(x^{\texttt{T}}_i, \widehat{m^{C_{\texttt{T}}\#}_{\phi_{C_{\texttt{T}}}}}\left(\eta^{C_{\texttt{T}}}_{ij}\right), \widehat{m^{N_{\texttt{T}}\#}_{\zeta_{N_{\texttt{T}}}}}\left(\eta^{N_{\texttt{T}}}_{ij}\right)\right)\right)\Bigg\|_2^2.
\end{aligned}
\end{equation}

We now make an important observation: although $\widehat{\mathcal{M}^{Y_{\texttt{T}}}_{\theta_{\texttt{T}}}}$ is trained using target domain data, when it is employed in the joint model, the distribution of its inputs is according to the source domain. This is a classical problem of covariate shift and can be easily tackled using state-of-the-art techniques that account for this shift (\cite{fatras2021jumbot, pmlr-v162-nguyen22d, pmlr-v238-manupriya24a}). The essential idea in such methods is to solve an OT problem between the domains and employ the obtained weights while computing the least square loss. This is typically written as a corrective loss term, $\ell_{\texttt{cs}}\left(\theta_{\texttt{T}},\phi_{\texttt{T}}, \phi_{\texttt{S}}, \zeta_{\texttt{T}}\right)$ (e.g., objective ($6$) in (\cite{10.1007/978-3-030-01225-0_28})). Thus, the overall loss for training the mechanisms on the target side is $\overline{\ell^{\texttt{T}}_{\texttt{gen}}}\left(\theta_{\texttt{T}},\phi_{\texttt{T}}, \phi_{\texttt{S}}, \zeta_{\texttt{T}}\right)\equiv\ell^{\texttt{T}}_{\texttt{gen}}\left(\theta_{\texttt{T}},\phi_{\texttt{T}},\zeta_{\texttt{T}}\right)+\ell_{\texttt{cs}}\left(\theta_{\texttt{T}},\phi_{\texttt{T}}, \phi_{\texttt{S}}, \zeta_{\texttt{T}}\right)$.% In our simulations we employ the JUMBOT framework for correcting this shift.

Finally, the effect-intrinsic, domain-intrinsic variables, $C,N$, need to be disambiguated correctly. This cannot be done if the training of the networks happen separately using the corresponding losses. For example, if identities of $C,N$ are interchanged, the domain-specific networks still remain the same, with an appropriate permutation of input variables. However, counterfactual inference may catastrophically fail if $C,N$ are interchanged. Hence, we propose to jointly train the NCMs with an additional loss term that helps disambiguate the variables. To this end, we assume a cost function, $\texttt{dis}$, is available that can perceive effect-intrinsic differences in the variables and ignores the domain-intrinsic differences. For example, in the case of image generation, the cost can be taken as the $L_{\texttt{pips}}$ loss \cite{zhang2018perceptual} or as squared Euclidean distance computed with image embeddings from a pre-trained network like AlexNet \cite{NIPS2012_c399862d}, VGG \cite{simonyan2015deepconvolutionalnetworkslargescale}, Swin-T \cite{9710580}, etc. Such a cost will be high if, say the identity of the object depicted in the images is not the same, and low even if, say the style of rendering of the images is different and vice-versa. Our novel loss term ($\ell_{\texttt{tr}}$) for disambiguating the variables is: 
\begin{equation}\label{eqn:Lt}
\begin{aligned}
\sum_{i=1}^{n _{\texttt{tr}}}\sum_{j=1}^{q_{\texttt{tr}}} \texttt{dis}\Bigg(\widehat{\mathcal{M}^{Y_{\texttt{S}}}_{\theta_{\texttt{S}}}}\left(x^{\texttt{S}}_i, \widehat{m^{C_{\texttt{S}}\#}_{\phi_{C_{\texttt{S}}}}}\left(\eta^{C_{\texttt{S}}}_{ij}\right), \widehat{m^{N_{\texttt{S}}\#}_{\zeta_{N_{\texttt{S}}}}}\left(\eta^{N_{\texttt{S}}}_{ij}\right)\right), \\\widehat{\mathcal{M}^{Y_{\texttt{T}}}_{\theta_{\texttt{T}}}}\left(x^{\texttt{S}}_i, \widehat{m^{C_{\texttt{S}}\#}_{\phi_{C_{\texttt{S}}}}}\left(\eta^{C_{\texttt{S}}}_{ij}\right), \widehat{m^{N_{\texttt{T}}\#}_{\zeta_{N_{\texttt{T}}}}}\left(\eta^{N_{\texttt{T}}}_{ij}\right)\right)\Bigg) = \ell_{\texttt{tr}} \cdot (n q)_{\texttt{tr}}
\end{aligned}
\end{equation}
Note that the effect-intrinsic variables are the same in the above in both the mechanisms, whereas the domain-intrinsic variables are different. Hence, for the correct $C,N$ pairs, this term will be low and vice-versa. In particular, if $C,N$ are interchanged, then this term will be high. Thus, this acts as a loss function to correctly disambiguate the exogenous variables. 
%The choice of the distance metric $d(\cdot, \cdot)$ can be chosen based on the downstream applications and particular datasets. For instance for standard regression for tabular data, Euclidean squared error may suffice, while for high-resolution images, an $L_2$ or $L_{\texttt{PIPS}}$ loss may need to be applied between embeddings of images derived from large pretrained models.
%Further note that $\overline{c_{ij}}$ don't need to be passed through $h$ but can be any values sampled from any distribution; for simplicity we can take samples from a Gaussian itself. The same holds for $x_i$, but for simplicity we can take those to be from the source dataset itself.\\
%This is connected to optimal transport in the sense that the cost function employed directly correlates to the cost function chosen to minimize the expected cost, as illustrated in Equation \ref{eqn:1}.

\subsection{Inference}
Given the learned joint NCM, counterfactual generation can be done using the standard three-step procedure: abduction, action, prediction. For the abduction step, we need samples from the posterior $p(c^{\texttt{S}}|x^{\texttt{S}},y^{\texttt{S}})$, where $(x^{\texttt{S}},y^{\texttt{S}})$ is the cause-effect pair observed in the source domain. To this end, we propose modeling the posterior, $p(c^{\texttt{S}},n^{\texttt{S}}|x^{\texttt{S}},y^{\texttt{S}})$, using a pushforward neural network generator, $g_{\psi}(x^{\texttt{S}}, y^{\texttt{S}}, N^{\#})$\footnote{$N^{\#}$ is the pushforward noise.}. From the Bayes rule, we have: $p(c^{\texttt{S}},n^{\texttt{S}}|x^{\texttt{S}},y^{\texttt{S}})p(x^{\texttt{S}},y^{\texttt{S}})=p(x^{\texttt{S}})p(c^{\texttt{S}})p(n^{\texttt{S}})p(y^{\texttt{S}}|x^{\texttt{S}},c^{\texttt{S}},n^{\texttt{S}})$, where the LHS is the factorization corresponding to the source causal graph and the RHS is the factorization involving the posterior to be estimated. Let $\mathcal{S}_{\theta_{\texttt{S}},\phi_{\texttt{S}},\zeta_{\texttt{S}}} \equiv \left(\Upsilon, \widehat{\mathcal{M}^{Y_{\texttt{S}}}_{\theta_{\texttt{S}}}}\left(\Upsilon\right)\right)$, be a set of samples from the source NCM, where $\Upsilon \equiv \left(x^{\texttt{S}}_i,\widehat{m^{C_{\texttt{S}}\#}_{\phi_{C_{\texttt{S}}}}}(\eta_{ij}^{C_{\texttt{S}}}),\widehat{m^{N_{\texttt{S}}\#}_{\zeta_{N_{\texttt{S}}}}}(\eta_{ij}^{N_{\texttt{S}}})\right)$ and $\eta^{{(C/N)}_{\texttt{S}}}_{ij}$ are independent samples of $\mathcal{N}(0,\texttt{I}_{d\times d})$. Let $\mathcal{P}_{\zeta_{N_{\texttt{S}}},\psi} \equiv \left(x^{\texttt{S}}_i,g_\psi\left(x^{\texttt{S}}_i,y^{\texttt{S}}_i,\eta_{ij}^{\#}\right),\widehat{m^{N_{\texttt{S}}\#}_{\zeta_{N_{\texttt{S}}}}}(\eta_{ij}^{N_{\texttt{S}}}),y^{\texttt{S}}_i\right)$, i.e. samples from the posterior, where $\eta^{N_{\texttt{S}}}_{ij},\eta^{\#}_{ij}$ are independent samples from $\mathcal{N}(0,\texttt{I}_{d\times d})$. With this notation, the loss for training $g_{\psi}$ turns out to be:
\begin{align}\label{eqn:pos_loss}
\ell_{\texttt{pos}}\left(\theta_{\texttt{S}},\phi_{\texttt{S}},\zeta_{\texttt{S}},\psi\right)\equiv \texttt{W}(\mathcal{S}_{\theta_{\texttt{S}},\phi_{\texttt{S}},\zeta_{\texttt{S}}},\mathcal{P}_{\zeta_{N_{\texttt{S}}},\psi}),
\end{align}
where, $\texttt{W}$ is an appropriate distance between between measures like the Wasserstein metric (\cite{NIPS2015_a9eb8122}). The training of the inference network can either be done jointly with the NCMs, in which case the optimization problem is:
\begin{align}\label{eqn:joint_final}
\min_{\theta_{\texttt{S}},\phi_{\texttt{S}},\zeta_{\texttt{S}},\theta_{\texttt{T}},\phi_{\texttt{T}},\zeta_{\texttt{T}},\psi}\ \ell^{\texttt{S}}_{\texttt{gen}} + \overline{\ell^{\texttt{T}}_{\texttt{gen}}} + \ell_{\texttt{tr}} + \ell_{\texttt{pos}}
\end{align}
or, can be trained separately in which case the optimization problems are:
\begin{equation}\label{eqn:ot}
\begin{aligned}[t]
(\theta_{\texttt{S}}^{*},\phi_{\texttt{S}}^{*},\zeta_{\texttt{S}}^{*},\theta_{\texttt{T}}^{*},\phi_{\texttt{T}}^{*},\zeta_{\texttt{T}}^{*})\equiv\textup{argmin}~\ell^{\texttt{S}}_{\texttt{gen}} + \overline{\ell^{\texttt{T}}_{\texttt{gen}}} + \ell_{\texttt{tr}}, \\
\psi^{*}\equiv\textup{argmin}_{\psi}\ \ell_{\texttt{pos}}\left(\theta_{\texttt{S}}^{*},\phi_{\texttt{S}}^{*},\gamma_{\texttt{S}}^{*},\psi\right)
\end{aligned}
\end{equation}
Once the optimal network parameters are obtained by solving the appropriate optimization problems using solvers like AdamW, counterfactual samples in the target domain can be produced by:
\begin{enumerate}
    \item Sampling from posterior using $g_{\psi_*}(x_s,y_s,N^\#)$. Different $n^\#$ samples from the reference Gaussian will produce different samples from the posterior.
    \item Counterfactual samples are obtained using $\widehat{\mathcal{M}^{Y_{\texttt{T}}}_{\theta^{*}_{\texttt{T}}}}\left(x_{\texttt{intv}}, g_{\psi}\left(x^{\texttt{S}}, y^{\texttt{S}},\eta^{\#}_{ij}\right), \widehat{m^{N_{\texttt{T}}\#}_{\zeta_{N_{\texttt{T}}}}}\left(\eta^{N_{\texttt{T}}}_{ij}\right)\right)$. Different $\eta^{\#/N_{\texttt{T}}}\sim\mathcal{N}(0,\texttt{I}_{d\times d})$ samples produce different samples from the counterfactual distribution.
\end{enumerate}
% \subsection{Connection to Optimal Transport}
\subsection{Connection to Conditional Optimal Transport}
Interestingly, in the special case where the distributions of the exogenous variables are the same in the source and the target domains\footnote{ $C_S\sim C_T$, $N_S\sim N_T$.}, (\ref{eqn:ot}) can be interpreted as a conditional OT problem. Let $p_{\theta_{\texttt{S}},\phi_{\texttt{S}}}(y^{\texttt{S}}|x^{\texttt{S}},\eta^{C_{\texttt{S}}}), p_{\theta_{\texttt{T}},\phi_{\texttt{S}}}(y^{\texttt{S}}|x^{\texttt{S}},\eta^{C_{\texttt{S}}})$ denote the distributions induced by the left \& right arguments to \texttt{dist} in (\ref{eqn:ot}) respectively. Furthermore, let $p_{\phi_{\texttt{S}}}(\eta)$ denote the distribution induced by $\widehat{m^{C_{\texttt{S}}\#}_{\phi_{C_{\texttt{S}}}}}$. Let $p_{\texttt{S}}(x^{\texttt{S}})$ denote the distribution of $X_{\texttt{S}}$. With this notation, it is easy to see that $\ell_{\texttt{tr}}$ is the sample mean corresponding to:
\vspace{-10px}
\begin{equation}
\begin{aligned}[t]
 \int\ p_{\theta_{\texttt{S}},\phi_{\texttt{S}}}(y^{\texttt{S}}|x^{\texttt{S}},\eta^{C_{\texttt{S}}}) p_{\theta_{\texttt{T}},\phi_{\texttt{S}}}(y^{\texttt{S}}|x^{\texttt{S}},\eta^{C_{\texttt{S}}})\texttt{d}(y^{\texttt{S}},y^{\texttt{T}})  \\ \int\int\ \Theta(x^{\texttt{S}}, \eta^{C_{\texttt{S}}}) p_{\phi_{\texttt{S}}}(\eta)p_{\texttt{S}}(x^{\texttt{S}}) \texttt{d}(n^{\texttt{C}_{\texttt{S}}}, x^{\texttt{S}})\\ =\mathbb{E}_{X_S,C_S}\left[\mathbb{E}\left[d(Y_S,Y^{'}_T)|X_S,C_S\right]\right],
\end{aligned}
\end{equation}\label{eqn:ot_theory}
where the total expectation is w.r.t. the joint causal model being learned\footnote{$\Theta(x^{\texttt{S}}, \eta^{C_{\texttt{S}}})$ is the first integral.}. Also, the source side marginal of this joint conditioned on $x^{\texttt{S}}$ is given by $\int\ p_{\theta_{\texttt{S}},\phi_{\texttt{S}}}(y^{\texttt{S}}|x^{\texttt{S}},\eta)p_{\phi_{\texttt{S}}}(\eta)\texttt{d}\eta$. The kernel least squares term in $\ell^{\texttt{S}}_{\texttt{gen}}$ matches this marginal to the distribution of $Y_S|x^{\texttt{S}}$ in the source dataset. Likewise, the $\ell^{\texttt{T}}_{\texttt{gen}}$ can be understood as a term for matching the target side conditional to $Y_T|x^{\texttt{S}}$. Moreover, if the distributions of the exogenous variables are the same, there is no covariate shift and hence $\overline{\ell^{\texttt{T}}_{\texttt{gen}}}$ is same as $\ell^{\texttt{T}}_{\texttt{gen}}$ in (\ref{eqn:ot}). In such a case, (\ref{eqn:ot}) becomes same as (6) in \cite{pmlr-v238-manupriya24a} (apart from the parametrization of the transport plan) and hence can be interpreted as a conditional OT problem between $p(y^{\texttt{S}}|x^{\texttt{S}})$ and $p(y^{\texttt{T}}|x^{\texttt{S}})$. Thus this connection seems to generalize the idea of employing OT based methods for applications like image-to-image translation \cite{korotin2023neural}, which is an example of generative modeling under domain shift (without any reference to causality). To summarize, in the special case where the domain shift is purely a conditional shift (change in causal mechanism) without any covariate shift, our model learned is essentially an OT plan corresponding to the conditional OT problem between $p(y^{\texttt{S}}|x^{\texttt{S}})$ and $p(y^{\texttt{T}}|x^{\texttt{S}})$. Nevertheless, even in this special case, our parametrization of the transport plan (joint) in terms of the effect-intrinsic variable $C_S$ is crucial for counterfactual generation.

\section{Simulations}
We present the details of the synthetic dataset on which we test our methodology. Due to the lack of an existing baseline tailored for this novel task, we introduce the following benchmark dataset that obeys the causal graph depicted in Figure \ref{fig:jointCausalGraph}, with structural equations that are described below.

\subsection{Dataset}
The SCMs of the source and target domains consist of $X, C, N$ variables that are of dimension $d\times1$. The corresponding effect variables in both domains i.e., $Y$ are of dimension $2d\times 1$, while their structural equations read as:
\begin{align}\label{eqn:source_mech}
    Y_{\texttt{S}} &= \begin{bmatrix}
\mathcal{A}_{\texttt{S}}e^{C_{\texttt{S}}}/2 \\
           (C_{\texttt{S}}\odot N_{\texttt{S}}) + X_{\texttt{S}}
         \end{bmatrix},
  \end{align}
\begin{align}\label{eqn:target_mech}
    Y_{\texttt{T}} &= \begin{bmatrix}
           (C_{\texttt{T}}\odot N^{2}_{\texttt{T}}) + X_{\texttt{T}} \\
\mathcal{A}_{\texttt{T}}e^{C_{\texttt{T}}}
         \end{bmatrix},
  \end{align}
where $\mathcal{A}_{\texttt{S/T}} = \mathcal{B}^{\texttt{T}}\mathcal{B} + 5I_{d\times d}$, and $\mathcal{B}_{\texttt{S/T}} = X_{\texttt{S/T}}(2^{X_{\texttt{S/T}}})^{\texttt{T}}$. In this context, the symbol $\odot$ denotes a Hadamard product, and '$\texttt{T}$' in the superscript refers to the transpose operation over a tensor. Likewise, the notation of the form $a^{X}$, $X^b$ for $a, b \in \mathbb{R}$ denote exponentiation to the base $a$, raising to the power of $b$ for each individual element of the vector $X \in \mathbb{R}^d$, respectively. The priors for the causal variables of the source \& target domains are:
\begin{equation}
\begin{aligned}\label{eqn:source_prior}
    X_{\texttt{S}} \sim \textbf{{\textrm{Uniform}}}(-1, 1), ~\widetilde{X_{\texttt{T}}} \sim \mathrm{\textbf{ContinuousBernoulli}}(0.6), \\~\widetilde{C_{\texttt{S}}} \sim \mathrm{\textbf{Beta}}(4, 5), ~C_{\texttt{T}} \sim \mathrm{\textbf{Normal}}(0, I_{d \times d}),\\ ~N_{\texttt{S}} \sim \mathrm{\textbf{vonMises}}(0,4), ~N_{\texttt{T}} \sim \mathrm{\textbf{Normal}}(0, 0.1\cdot I_{d \times d}).
\end{aligned}
\end{equation}
Infact, we pick $C_{\texttt{S}}, X_{\texttt{T}}$ to be an affine transforms of $\widetilde{C_{\texttt{S}}}, \widetilde{X_{\texttt{T}}}$ respectively, as $C_{\texttt{S}} = 1_{d \times 1} - (2\cdot\widetilde{C_{\texttt{S}}})$, $X_{\texttt{T}} = 1_{d \times 1} - (2\cdot\widetilde{X_{\texttt{T}}})$.
For the target domain, we explicitly choose different priors so as to establish a strong benchmark that does not utilize spurious correlations across variables and domains.

\subsubsection{Computing Shifted Counterfactuals}\label{subsubsec:css}
A key property of the proposed dataset is that the source-side mechanism is invertible, given the values of $X_\texttt{S}, Y_{\texttt{S}}$. This is in line with invertibility assumptions often made in the SCM community \cite{pmlr-v202-nasr-esfahany23a,ijcai2024p907}. Let $Y_{\texttt{S}}[\texttt{:d}],Y_{\texttt{S}}[\texttt{d:}]$ denote the first and last $d$ elements of $Y_{\texttt{S}} \in \mathbb{R}^{2d}$ respectively. Then, given $X_\texttt{S}, Y_{\texttt{S}}$, we can compute $C_\texttt{S} = \mathcal{A}^{-1}_{\texttt{S}}Y_{\texttt{S}}[\texttt{:d}]$. Notice that by design, $\mathcal{A}_{\texttt{S/T}}$ is always positive definite and hence invertible, so such an inverse is always defined. Using this $C_{\texttt{S}}$ we can then compute $N_{\texttt{S}} = (Y_{\texttt{S}}[\texttt{d:}] - X_{\texttt{S}})/C_{\texttt{S}}$. Thus for each triplet $(x_{\texttt{intv}}, x_{\texttt{fact}}, y_{\texttt{fact}})$, there exists a unique counterfactual for that in the source domain. 
However, this is not the case in the target domain. Thought $C_{\texttt{T}}$ is exactly retrievable given $X_{\texttt{T}}, Y_{\texttt{T}}$, $N_{\texttt{T}}$ is not, since there are at least two solutions for it given $X_{\texttt{T}}, Y_{\texttt{T}}$ and $C_{\texttt{T}}$, namely $N_{\texttt{T}} = \pm \sqrt{(Y_{\texttt{T}}[\texttt{d:}] - X_{\texttt{T}})/C_{\texttt{T}}}$. This illustrates that the distribution of shifted counterfactuals is crucially dependent on the target side mechanism. Thus, to generate shifted counterfactual groundtruth samples, we follow the procedure outlined below:
\begin{enumerate}
    \item Sample $x^{\texttt{T}}_{\texttt{intv}}$ for the target domain from any distribution with identical support as the prior of $X_{\texttt{T}}$. For simplicity, we may choose this distribution to be the prior of $X_{\texttt{T}}$ itself.
    \item Sample $x^{\texttt{S}}_{\texttt{fact}}, c^{\texttt{S}}, n^{\texttt{S}}$ from the priors of $X_{\texttt{S}}, C_{\texttt{S}}, N_{\texttt{S}}$ respectively. Generate the corresponding $y^{\texttt{S}}_{\texttt{fact}}$ using Equation \ref{eqn:source_mech}. Accordingly we randomly select a pair $(x^{\texttt{S}}_{\texttt{fact}}, y^{\texttt{S}}_{\texttt{fact}})$.
    \item Using the invertibility of Equation \ref{eqn:source_mech}, compute $c^{\texttt{S}}_{\texttt{fact}}$ for the chosen $(x^{\texttt{S}}_{\texttt{fact}}, y^{\texttt{S}}_{\texttt{fact}})$ pair.
    \item Sample $n^{\texttt{T}}$ from the prior of $N_{\texttt{T}}$ and then plug $x^{\texttt{T}}_{\texttt{intv}}$, $c^{\texttt{S}}_{\texttt{fact}}$, $n^{\texttt{T}}$ into Equation \ref{eqn:target_mech} to generate samples from the distribution of shifted counterfactuals.
\end{enumerate}
The distribution of shifted counterfactuals considered above is as conditioned on a particular triplet $(x^{\texttt{T}}_{\texttt{intv}}, x^{\texttt{S}}_{\texttt{fact}}, y^{\texttt{S}}_{\texttt{fact}})$. Sampling multiple such triplets and generating corresponding shifted counterfactuals then corresponds to generating joint samples of shifted counterfactuals paired with their causal variables of interest.

\subsection{Results}
We test our proposed methodology on the above specified dataset with $d=1$, as this is sufficient to demonstrate the generation of shifted counterfactuals. 
When training on this dataset as per Equation \ref{eqn:joint_final}, we simply instantiate the following MSE loss for $\ell_{\texttt{tr}}$ (cf. Equation \ref{eqn:Lt}):
\begin{equation}
\begin{aligned}\label{eqn:syn_Lt}
\frac{1}{mq}\sum_{i=1}^m\sum_{j=1}^q \Bigg \|\widehat{\mathcal{M}^{Y_{\texttt{S}}}_{\theta_{\texttt{S}}}}\Big(\cdot,\phi_{\texttt{S}},\zeta_{\texttt{S}}\Big)[0]- \frac{1}{2}\widehat{\mathcal{M}^{Y_{\texttt{S}}}_{\theta_{\texttt{T}}}}\left(\cdot,\phi_{\texttt{S}},\zeta_{\texttt{T}}\right)[1]\Bigg\|_2^{2}.
\end{aligned}
\end{equation}
This choice nicely fits in with the intention of disambiguating the latent context from the noise across domains. To see this, notice that the synthetic dataset proposed above bears the property that if $X_{\texttt{S}} = X_{\texttt{T}}$ and $C_{\texttt{S}} = C_{\texttt{T}}$, then $Y_{\texttt{S}} = \frac{Y_{\texttt{T}}}{2}$; thus $C_{\texttt{S/T}}$ satisfies an invariant across domains which $N_{\texttt{S/T}}$ does NOT satisfy, which in turn helps disentangle the role of $C_{\texttt{S}}$ when performing shifted counterfactual inference.

To evaluate the quality of our shifted counterfactuals, we run a two sample test \cite{schrab2021mmd} between samples from the estimated joint distribution of factuals and shifted counterfactuals, and the corresponding groundtruth distribution. In particular, given a pair $(x_i^{\texttt{S}}, x_{\texttt{intv}_i})$, we randomly sample $100$ triplets $\left(\widehat{m^{C_{\texttt{S}}\#}_{\phi_{C_{\texttt{S}}}}}\left(\eta^{C_{\texttt{S}}}_{ij}\right), \widehat{m^{N_{\texttt{S}}\#}_{\zeta_{N_{\texttt{S}}}}}\left(\eta^{N_{\texttt{S}}}_{ij}\right),\widehat{m^{N_{\texttt{T}}\#}_{\zeta_{N_{\texttt{T}}}}}\left(\eta^{N_{\texttt{T}}}_{ij}\right)\right)$ and compare samples from $\left[\widehat{y^{\texttt{S}}_{ij}};~\widehat{\mathcal{M}^{Y_{\texttt{T}}}_{\widehat{\theta}_{\texttt{T}}}}\left(x_{\texttt{intv}_i}, g_\psi\left(x^{\texttt{S}}_i,\widehat{y^{\texttt{S}}_{ij}},\eta_{ij}^{\#}\right),\widehat{m^{N_{\texttt{T}}\#}_{\zeta_{N_{\texttt{T}}}}}\left(\eta_{ij}^{N_{\texttt{T}}}\right)\right)\right]$ and $\left[y^{\texttt{S}}_{ij};~\mathcal{M}^{Y_{\texttt{T}}*}\left(x_{\texttt{intv}_i}, c_{\texttt{S}|(x^{\texttt{S}}_i,y^{\texttt{S}}_{ij})},N_{\texttt{T}}\right)\right]$. Here $y^{\texttt{S}}_{ij}/ \widehat{y^{\texttt{S}}_{ij}}$ are sampled by plugging in $x^{\texttt{S}}_i$ in the estimated /groundtruth equations and varying the context and noise from the source domain appropriately. Finally, we average the Agg-MMD two-sample test scores between samples of these distributions over $500$ pairs of $(x_i^{\texttt{S}}, x_{\texttt{intv}_i})$ and got a $70\%$ score with $95\%$ confidence, proving the efficacy of our approach\footnote{We will provide code upon acceptance.}.

We adopt the two-stage training process as explained in Equations \ref{eqn:ot}. This is done to ensure that the posterior network learns to be consistent with the distribution of samples induced from the trained generator network. The hyperparameters for controlling individual loss terms are to be picked via cross-validation. 
Details on the architectures of the generator, context and noise networks are provided in the Appendix \ref{subsec:appendix:training}.
% We pick the generator architectures to be simple three-layered MLPs with $32$ hidden units in each, with PReLU (\cite{7410480}) activation. The same holds for the posterior network architecture as well. The architectures for context and noise networks are simply one-layered MLPs with $32$ hidden units and PReLU activation.

\section{Summary and Conclusions}
In this work, we introduced the novel task of structural-counterfactual generation under general domain-shift and studied it in the practical setting where no parallel/joint data is available. It turned out that modeling and training are considerably more challenging than in classical counterfactual generation tasks. In terms of modeling, the crucial step was to separately model the effect-intrinsic and domain-intrinsic variables, so that the structural-counterfactual is well-defined. In terms of training, disambiguating the effect-intrinsic and domain-intrinsic exogenous variables to correctly learn their distributions was the challenging step. From the trained model, the procedure for sampling counterfactuals in the target domain was presented. Connections to the optimal transport problem provided more insight into the proposed methodology.

In future, we would like to explore vision, biology applications of the novel task studied in this paper. Using the connections to OT and corresponding barycenter problems, we plan to extend structural-counterfactual generation to settings of domain interpolation.

% \section*{Accessibility}
% Authors are kindly asked to make their submissions as accessible as possible for everyone including people with disabilities and sensory or neurological differences.
% Tips of how to achieve this and what to pay attention to will be provided on the conference website \url{http://icml.cc/}.

% \section*{Software and Data}

% If a paper is accepted, we strongly encourage the publication of software and data with the
% camera-ready version of the paper whenever appropriate. This can be
% done by including a URL in the camera-ready copy. However, \textbf{do not}
% include URLs that reveal your institution or identity in your
% submission for review. Instead, provide an anonymous URL or upload
% the material as ``Supplementary Material'' into the OpenReview reviewing
% system. Note that reviewers are not required to look at this material
% when writing their review.

% Acknowledgements should only appear in the accepted version.
% \section*{Acknowledgements}

% \textbf{Do not} include acknowledgements in the initial version of
% the paper submitted for blind review.

% If a paper is accepted, the final camera-ready version can (and
% usually should) include acknowledgements.  Such acknowledgements
% should be placed at the end of the section, in an unnumbered section
% that does not count towards the paper page limit. Typically, this will 
% include thanks to reviewers who gave useful comments, to colleagues 
% who contributed to the ideas, and to funding agencies and corporate 
% sponsors that provided financial support.

\section*{Impact Statement}

This paper presents work whose goal is to advance the field of 
Machine Learning. There are many potential societal consequences 
of our work, none which we feel must be specifically highlighted here.

% In the unusual situation where you want a paper to appear in the
% references without citing it in the main text, use \nocite
% \nocite{langley00}

\bibliography{main}
\bibliographystyle{icml2025}

%%%%%%%%%%%%%%%%%%%%%%%%%%%%%%%%%%%%%%%%%%%%%%%%%%%%%%%%%%%%%%%%%%%%%%%%%%%%%%%
%%%%%%%%%%%%%%%%%%%%%%%%%%%%%%%%%%%%%%%%%%%%%%%%%%%%%%%%%%%%%%%%%%%%%%%%%%%%%%%
% APPENDIX
%%%%%%%%%%%%%%%%%%%%%%%%%%%%%%%%%%%%%%%%%%%%%%%%%%%%%%%%%%%%%%%%%%%%%%%%%%%%%%%
%%%%%%%%%%%%%%%%%%%%%%%%%%%%%%%%%%%%%%%%%%%%%%%%%%%%%%%%%%%%%%%%%%%%%%%%%%%%%%%
\newpage
\appendix
\onecolumn
\section{Appendix}\label{sec:appendix}

% \subsection{Prolegomena}\label{subsec:appendix:pre}

% \subsubsection{Causality}\label{subsubsec:appendix:causality}
% We reproduce notation, terminology and quintessential concepts of causality particularly in relation to $\mathcal{L}_{1/2/3}$ distributions and counterfactuals from \cite{10.1145/3501714.3501743}, for completeness and convenience for the interested reader. 

% % \textbf{Definition 7} (Layer 3 Valuation)\textbf{.} An SCM \emph{M} =
% % \emph{⟨}\textbf{U}\emph{,}\textbf{V}\emph{,F
% % ,P}(\textbf{U})\emph{⟩}induces a family of joint distributions over
% % counterfactual events \textbf{Yx}\emph{,...,}\textbf{Zw}, for any
% % \textbf{Y}\emph{,}\textbf{Z}\emph{,...,}\textbf{X}\emph{,}\textbf{W}
% % \emph{⊆}\textbf{V}:\\
% % \emph{PM}(\textbf{yx}\emph{,...,}\textbf{zw}) = ∑
% % \emph{P}(\textbf{u})\emph{.} (1.16) \emph{\{}\textbf{u}
% % \emph{\textbar{}} \textbf{Yx}(\textbf{u})=\textbf{y}\emph{,}\\
% % \emph{...,} \textbf{Zw}(\textbf{u})=\textbf{z}\emph{\}}\\

% \subsubsection{Optimal Transport}\label{subsubsec:appendix:ot}

\subsection{Training}\label{subsec:appendix:training}
We use symmetric design for our work i.e. we use same architecture for both source side networks and target side networks. We use $5$ hidden layers for $f_\theta$. For context network $h_\phi$, we use single hidden layer. For our noise networks, $t_\gamma$, we use $3$ hidden layers. For every hidden layer, hidden dimension is $128$ and every linear layer(except the last layer) is followed by PReLU \cite{7410480}. PReLU is initialized with $0.25$ for $f_\theta$, and with $0.01$ for both the context and noise networks. We used skip-connections in the context network and noise network, since we observed better empirical performance using them.

We train with a dataset size of $25000$. We use AdamW optimizer in stage $1$ training and Adam in stage $2$ training. We use linear increase followed by cosine decrease learning rate scheduler. We set the kernel hyperparameter $\varrho$ to $1.0$ throughout.
% \subsection{Evaluation}\label{subsec:appendix:eval}

%%%%%%%%%%%%%%%%%%%%%%%%%%%%%%%%%%%%%%%%%%%%%%%%%%%%%%%%%%%%%%%%%%%%%%%%%%%%%%%
%%%%%%%%%%%%%%%%%%%%%%%%%%%%%%%%%%%%%%%%%%%%%%%%%%%%%%%%%%%%%%%%%%%%%%%%%%%%%%%

\end{document}